\documentclass[a4paper,runningheads]{llncs}

\usepackage[T1]{fontenc}

\usepackage{graphicx}
\usepackage[hidelinks]{hyperref}
\usepackage{color}

\usepackage{subcaption}
\usepackage{array}

\usepackage{scalerel}
\usepackage{tikz}
\usetikzlibrary{svg.path}

\definecolor{orcidlogocol}{HTML}{A6CE39}
\tikzset{
  orcidlogo/.pic={
    \fill[orcidlogocol] svg{M256,128c0,70.7-57.3,128-128,128C57.3,256,0,198.7,0,128C0,57.3,57.3,0,128,0C198.7,0,256,57.3,256,128z};
    \fill[white] svg{M86.3,186.2H70.9V79.1h15.4v48.4V186.2z}
                 svg{M108.9,79.1h41.6c39.6,0,57,28.3,57,53.6c0,27.5-21.5,53.6-56.8,53.6h-41.8V79.1z M124.3,172.4h24.5c34.9,0,42.9-26.5,42.9-39.7c0-21.5-13.7-39.7-43.7-39.7h-23.7V172.4z}
                 svg{M88.7,56.8c0,5.5-4.5,10.1-10.1,10.1c-5.6,0-10.1-4.6-10.1-10.1c0-5.6,4.5-10.1,10.1-10.1C84.2,46.7,88.7,51.3,88.7,56.8z};
  }
}

\newcommand\orcidicon[1]{\href{https://orcid.org/#1}{\mbox{\scalerel*{
\begin{tikzpicture}[yscale=-1,transform shape]
\pic{orcidlogo};
\end{tikzpicture}
}{|}}}}

\newcommand{\mytitle}{ThermoCycleNet: Stereo-based Thermogram Labeling for Model Transition to Cycling}

\hypersetup{
    pdftitle={\mytitle},    
    pdfsubject={Abstract},
    pdfauthor={D. Andrés López et al.},    
    pdfkeywords={Automatic Labeling, Deep Neural Network, Infrared Thermography, Multimodal Stereo Transformation, Physical Exercise Testing, Semantic Segmentation}, 
    plainpages=false,           
    colorlinks=false,           
    pdfborder={0 0 0},          
    breaklinks=true,            
    pdfpagemode=UseOutlines
}

\usepackage[
type={CC},
modifier={by},
version={4.0},
]{doclicense}

\usepackage[]{eso-pic}

\begin{document}

\title{\mytitle\thanks{Funding: Johannes Gutenberg University ``Stufe I'': ``Start ThermoCycleNet''. \\Partial funding: Carl-Zeiss-Stiftung: ``Multi-dimensionAI'' (CZS-Project number: P2022-08-010). } \thanks{Presented at IWANN 2025 18th International Work-Conference on Artificial Neural Networks, A Coruña, Spain, 16-18 June, 2025. Book of abstracts: ISBN: 979-13-8752213-1}}
\titlerunning{ThermoCycleNet}
\subtitle{Extended Abstract}

\author{
Daniel Andrés López\inst{1}\orcidID{\orcidicon{0000-0001-9045-7642}}
\and Vincent Weber\inst{2}\orcidID{\orcidicon{0000-0001-9138-4932}}
\and Severin Zentgraf\inst{2}\orcidID{\orcidicon{0009-0009-7217-0491}}
\and \\Barlo Hillen\inst{2,3}\orcidID{\orcidicon{0000-0002-5090-2286}}
\and Perikles Simon\inst{2}\orcidID{\orcidicon{0000-0002-7996-4034}}
\and Elmar Schömer\inst{1}\orcidID{\orcidicon{0000-0002-5652-2591}}
}

\authorrunning{D. Andrés López et al.}

\institute{
Institute of Computer Science, Research Group Computational Geometry, Johannes Gutenberg University, Mainz, Germany \\ \email{\{daniel.andres, schoemer\}@uni-mainz.de}
\and
Institute of Sports Science, Department of Sports Medicine, Disease Prevention and Rehabilitation, Johannes Gutenberg University, Mainz, Germany \\ \email{\{v.weber, s.zentgraf, b.hillen, simonpe\}@uni-mainz.de}
\and
Institute of Occupational, Social and Environmental Medicine, University Medical Center, Mainz, Germany
}

\maketitle
\makebox[\linewidth]{\small August 1, 2025}

\begin{abstract}Infrared thermography is emerging as a powerful tool in sports medicine, allowing assessment of thermal radiation during exercise and analysis of anatomical regions of interest, such as the well-exposed calves. Building on our previous advanced automatic annotation method, we aimed to transfer the stereo- and multimodal-based labeling approach from treadmill running to ergometer cycling. Therefore, the training of the semantic segmentation network with automatic labels and fine-tuning on high-quality manually annotated images has been examined and compared in different data set combinations. The results indicate that fine-tuning with a small fraction of manual data is sufficient to improve the overall performance of the deep neural network. Finally, combining automatically generated labels with small manually annotated data sets accelerates the adaptation of deep neural networks to new use cases, such as the transition from treadmill to bicycle.

\footnotesize
\keywords{Automatic Labeling \and Deep Neural Network \and Infrared Thermography \and Multimodal Stereo Transformation \and Physical Exercise Testing \and Semantic \mbox{Segmentation}.}
\end{abstract}
\normalsize

\AddToShipoutPictureBG*{\AtPageLowerLeft{%
    \put(\LenToUnit{4.2cm},\LenToUnit{4.2cm}){%
        \href{https://creativecommons.org/licenses/by-nc/4.0/}{\ccby} \href{https://doi.org/10.5281/zenodo.16675334}{10.5281/zenodo.16675334} }
    }
  }

\clearpage
\paragraph{Introduction:}
Infrared thermography (IRT) has been applied to several use cases in sports and medicine such as exercise testing in treadmill running and pedaling on a bicycle ergometer~\cite{Hillen2023}. Previously, we developed a stereo system for label generation consisting of an infrared, a time-of-flight, and a visual camera~\cite{AndresLopez2024} for treadmill running. The supervised learning was employed based on labels created automatically in the visual domain and transformed to the IRT domain with stereo properties and known depth information. However, the extension of a specialized trained deep neural network to an additional use case is a challenging. The aim of this study was to adapt an advanced stereo label generation method~\cite{AndresLopez2024} for ergometer cycling and simultaneously develop a new model to sufficiently segment moving legs in running and cycling. We hypothesize that automatically stereo-generated labels enable a fast deep neural network transition from running to cycling and only a few manual labeled data are enough to fine-tune the network.

\paragraph{Methods:}
The previous StereoThermoLegs~\cite{AndresLopez2024} data set was tested in conjunction with a manual data set (training set with 670 images and 200 test images) to evaluate if pre-training with automatic labels and fine-tuning with manual labels sufficiently segment the targeted regions of interest. Thereby, five models were analyzed on a manually labeled test set: Model M trained only with manual data, and models S1-S4 with stereo pre-training and different fractions of the manual data set (0\%, 10\%, 50\%, 100\%) for fine-tuning. Based on new label generation models in the visual domain, the stereo labeling method has been transferred from running to cycling. Additionally, a new manual data set was labeled. Subsequently, the new ThermoCycleNet was trained on all of these stereo-labeled data and fine-tuned on the manual data sets similar to~\cite{AndresLopez2024}.

\paragraph{Results:}
Network M immediately performed sufficiently (Table~\ref{tab:results_iou}). Model S1 showed inferior performance on manual test data, but S2 significantly surpassed S1. Only S3 and S4 outperform baseline M in overall and target class performance.

\begin{table}[ht]
    \centering
    \setlength{\belowcaptionskip}{-15pt}
    \begin{tabular}{p{1.8cm}||>{\centering\arraybackslash}p{1.9cm}|>{\centering\arraybackslash}p{1.9cm}|>{\centering\arraybackslash}p{1.9cm}|>{\centering\arraybackslash}p{1.9cm}|>{\centering\arraybackslash}p{1.9cm}}
    \textbf{Network}  & \textbf{M}      & \textbf{S1}     & \textbf{S2}     & \textbf{S3}     & \textbf{S4}              \\ \hline
    Pre-training       & manual          & stereo          & stereo          & stereo          & stereo                   \\
    Fine-tuning       & -               & -               & 10\% manual     & 50\% manual     & 100\% manual             \\ \hline
    \textit{Mean IoU} & \textit{0.6752} & \textit{0.5067} & \textit{0.6076} & \textit{0.7088} & \textit{\textbf{0.7166}} \\ \hline
    Left calf    & 0.8255          & 0.7315          & 0.7846          & 0.8440          & \textbf{0.8650}          \\ \hline
    Right calf   & 0.8619          & 0.6231          & 0.7833          & 0.8752          & \textbf{0.8806}
    \end{tabular}
    \caption{Results of the Intersection over Union (IoU) for the semantic segmentation of calves with different combinations of pre-training and fine-tuning data sets.}
    \label{tab:results_iou}
\end{table}

\begin{figure}[htb]
    \centering
    \begin{subfigure}[t]{0.23\textwidth}
        \centering
        \begin{minipage}{\textwidth}
            (a) \includegraphics[width=0.6\textwidth]{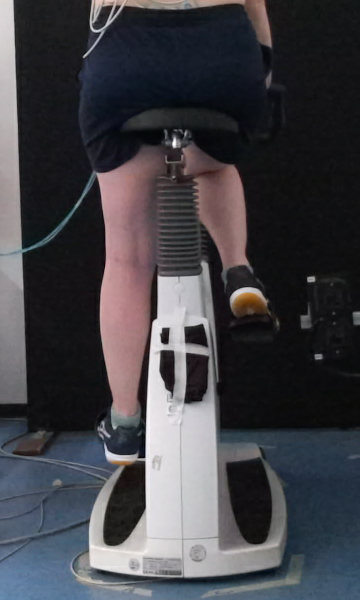}
        \end{minipage}
    \end{subfigure}
    \hfill
    \begin{subfigure}[t]{0.23\textwidth}
        \centering
        \begin{minipage}{\textwidth}
        (b) \includegraphics[width=0.6\textwidth]{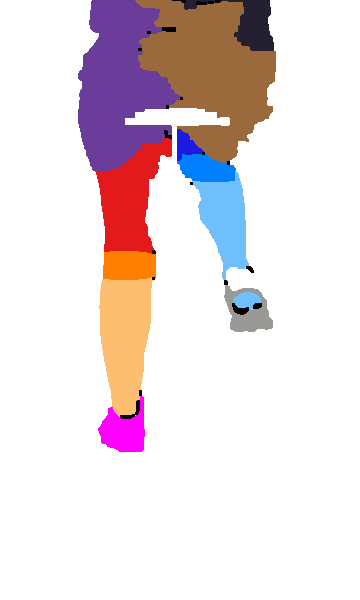}
        \end{minipage}
    \end{subfigure}
    \hfill
    \begin{subfigure}[t]{0.23\textwidth}
        \centering
        \begin{minipage}{\textwidth}
        (c)    \includegraphics[width=0.6\textwidth]{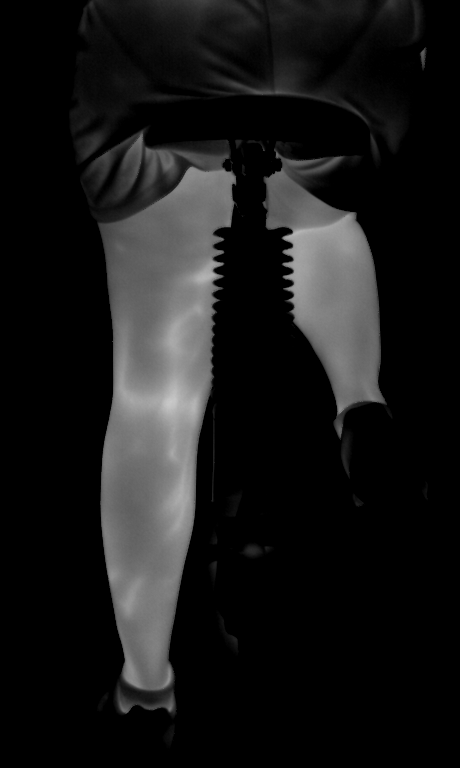}
        \end{minipage}
    \end{subfigure}
    \hfill
    \begin{subfigure}[t]{0.23\textwidth}
        \centering
        \begin{minipage}{\textwidth}
        (d)    \includegraphics[width=0.6\textwidth]{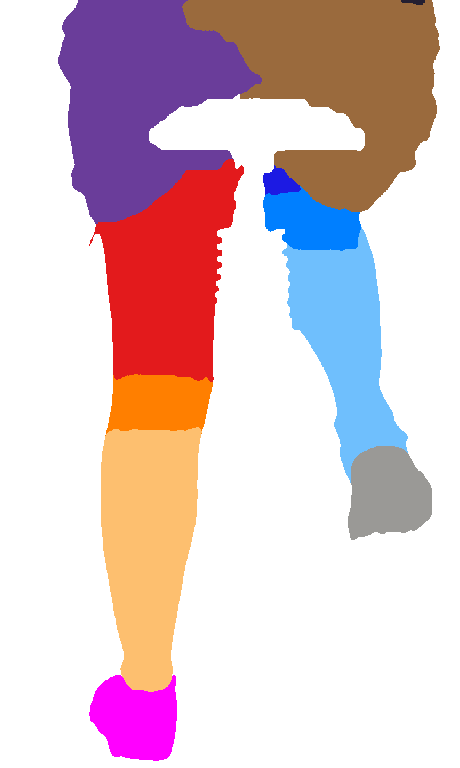}
        \end{minipage}
    \end{subfigure}
    \setlength{\belowcaptionskip}{-30pt}
    \caption{Example of an automatic stereo label for ThermoCycleNet: visual image (a), generated label (b), thermogram (c), and transformed label (d).\newline\newline}
    \label{fig:stereocyclenet-example}
\end{figure}

The stereo-labeled cycling data set has 1600 thermograms from a single case (see Figure~\ref{fig:stereocyclenet-example}) and 560 manually annotated training and 120 test images.
ThermoCycleNet achieves a mean IoU of 0.51 on the new data set, the target classes of the left and right calf are 0.75 and 0.72 respectively. 

\paragraph{Conclusions:}
Incorporating a high quantity of low-quality stereo labels enables a fast transition from running to cycling. However, only fine-tuning with high-quality manual labels leads to high model performance in line with \cite{Gunes2023a}. Strikingly, many automatic labels combined with a few manual labels are sufficient for a high-performance ThermoCycleNet to segment the calves of runners and cyclists.

\begin{credits}
\subsubsection{\ackname}
We would like to thank all participants for their voluntary participation in this study. Written informed consent was obtained from all participants.
\subsubsection{\discintname}
The authors have no competing interests to declare that are relevant to the content of this article.
\end{credits}

\bibliographystyle{splncs04}
\bibliography{stereocyclenet}

\end{document}